\documentclass{article}
\usepackage[utf8]{inputenc}
\usepackage{geometry}
\usepackage{mathptmx} 
\usepackage{amsmath}
\usepackage{graphicx}
\usepackage{booktabs}
\usepackage{hyperref}
\usepackage{tikz}
\usetikzlibrary{shapes, arrows.meta, positioning, calc, fit, backgrounds, shadows}

\tikzset{
    basic/.style={draw=black!60, thick, rounded corners=3pt, align=center, font=\small},
    user/.style={basic, circle, fill=gray!10, minimum size=1.2cm},
    agent/.style={basic, rectangle, fill=blue!5, minimum width=2.5cm, minimum height=1cm},
    dynamic/.style={basic, rectangle, fill=orange!10, dashed, minimum width=2.5cm, minimum height=1cm},
    state/.style={basic, rectangle, fill=green!5, align=left, font=\scriptsize, minimum width=3cm, minimum height=1.5cm},
    arrow/.style={-Latex, thick},
    query/.style={-Latex, dashed, red, thick},
    feedback/.style={-Latex, dotted, blue, thick}
}

\title{\textbf{IACT: A Self-Organizing Recursive Model for General AI Agents} \\ \large A Technical White Paper on the Architecture Behind kragent.ai}
\author{Pengju Lu \\ \texttt{pjlu@mail.bnu.edu.cn}}
\date{November 2025}

\begin{document}

\maketitle

\let\thefootnote\relax\footnotetext{The prototype (AIlice) was open-sourced in 2023 at https://github.com/myshell-ai/AIlice. The current system is available at https://kragent.ai.}

\begin{abstract}This technical white paper introduces the \textbf{Interactive Agents Call Tree (IACT)}, a computational model designed to address the limitations of static, hard-coded agent workflows. Unlike traditional systems that require pre-defined graphs or specialized programming, IACT operates as a \textbf{general-purpose autonomous system driven purely by user dialogue}.

   Given a high-level objective, the system \textbf{autonomously grows a dynamic, recursive agent topology incrementally tailored to the problem's structure}. This allows it to scale its organizational complexity to match open-ended tasks. To mitigate the error propagation inherent in unidirectional function calls, IACT introduces interactional redundancy by replacing rigid invocations with \textbf{bidirectional, stateful dialogues}. This mechanism enables runtime error correction and ambiguity resolution. We describe the architecture, design principles, and practical lessons behind the production deployment of this model in the \textit{kragent.ai} system, presenting qualitative evidence from real-world workflows rather than exhaustive benchmark results.\end{abstract}

\vspace{1cm}

\section{Introduction}

The transition from statistical pattern matching to large-scale generative reasoning marks a paradigm shift in computer science. In this new era, the Large Language Model (LLM) functions analogously to a probabilistic \textbf{Central Processing Unit (CPU)}: it accepts an instruction and context (Input), performs reasoning operations, and generates a sequence of executable tokens (Output). However, a CPU alone does not constitute a complete computing system. To address complex, non-trivial problems—ranging from software engineering to scientific research—raw processing power must be organized within a robust architecture that manages memory, Input/Output (I/O) protocols, and execution flow.

Current approaches to building LLM-based applications often lack a rigorous structural foundation. One prevalent paradigm, \textbf{Chain-of-Thought (CoT) workflows \cite{wei2022chain}} (e.g., ReAct loops \cite{yao2023react}), relies on a single linear thread of execution. While effective for short tasks, these systems suffer from \textit{error cascading}: a single logical fallacy early in the chain often leads to irreversible failure. Furthermore, as the conversation history grows, the context window becomes saturated, degrading the model's reasoning performance. A second paradigm, \textbf{Multi-Agent Group Chat}, attempts to solve this by allowing multiple agents to converse freely. However, without structural constraints, these systems often devolve into chaotic "noise," where agents struggle to maintain focus due to context pollution from irrelevant peers.

Current approaches, particularly static Directed Acyclic Graphs (DAGs) or fixed Standard Operating Procedures (SOPs), suffer from a fundamental constraint: \textbf{The inflexibility of static topologies}. They operate on the assumption that the complexity of a solution path can be predetermined by the developer. However, in open-ended domains, task complexity is dynamic and often emergent. A static graph inevitably fails when a task deviates from its pre-wired logic.

To address this, we propose the \textbf{Interactive Agents Call Tree (IACT)}. Unlike static graphs, IACT enables a \textbf{dynamic, runtime-adaptive topology} that emerges incrementally during execution. Crucially, to mitigate the \textbf{Error Propagation} inherent in unidirectional function calls, IACT replaces rigid invocations with \textbf{Bidirectional, Stateful Dialogues}. This mechanism enables continuous runtime verification and correction, allowing the system to scale complexity without sacrificing reliability.

In this paper, we detail the architecture of kragent.ai based on this model. \textbf{This work serves as a system architecture description and experience report from production environments.} Unlike traditional research papers focused on quantitative benchmarks, we present \textbf{design principles and qualitative evidence} drawn from real-world workflows.

\section{Design Philosophy}

The architecture of IACT is grounded in two fundamental principles that represent a departure from traditional "Workflow Engineering" towards a native "Agentic Computing" paradigm.

\subsection{Dynamic Context Construction}
   IACT minimizes the reliance on static system prompts, instead adopting a strategy of \textbf{Dynamic Instruction Injection}. Rather than encoding all behavioral guidelines into a fixed instruction set, the system injects specific prompts dynamically into the context as events occur. This approach allows the agent's trajectory to be flexibly steered by environmental feedback.

   For instance:
   \begin{itemize}
       \item \textbf{Resource Monitoring}: The system monitors context usage and injects a "Context Overflow" warning only when a threshold is breached, prompting the agent to initiate compression.
       \item \textbf{Environmental Grounding}: To mitigate hallucinations regarding the file system, the system actively injects the current working directory and file lists, grounding the agent in the actual environment.
       \item \textbf{Exception Handling}: Runtime errors are not suppressed but injected as raw stack traces. This guides the agent to treat exceptions as standard inputs to be analyzed and resolved.
   \end{itemize}

   By shifting behavioral guidance from the static system prompt to dynamic injections, IACT reduces the cognitive load on the model and transforms the context window into a \textbf{live information dashboard}, ensuring execution fluidity. Crucially, this dynamism is architected to preserve inference efficiency. Despite frequent updates, the context structure remains compatible with KV Cache Optimization (detailed in Section 4.4).

\subsection{LLM-Centric Control Flow}
   A central debate in agent system design is the locus of control: should the workflow be defined by the human developer or by the model? IACT adopts a strictly \textbf{LLM-Centric Control Flow}. We argue that \textbf{intelligence cannot be explicitly scheduled}; hard-coded workflows are brittle and fail when tasks deviate from pre-defined paths.

   In IACT, "Control Flow" is defined as the autonomous decision-making process across four key dimensions:
   \begin{itemize}
       \item \textbf{Agent Instantiation}: Deciding which specialized agents to create to address specific sub-problems.
       \item \textbf{Agent Invocation}: Determining the target, timing, and content of calls to child agents.
       \item \textbf{Vertical Interaction}: Deciding the timing and content of communication with the parent agent.
       \item \textbf{Tool Execution}: Selecting and executing tools to interact with the environment.
   \end{itemize}

   The system provides the "what"—the tools, the communication protocols, and the high-level principles—but delegates the "how" entirely to the agent.

   This philosophy manifests as \textbf{Adaptive Topology}: the organizational structure is not pre-wired but \textbf{grows} organically based on the LLM's real-time judgment. For example, when a Coder agent encounters an unfamiliar library during implementation, it can spontaneously instantiate a Researcher agent to retrieve documentation and examples, seamlessly expanding the topology to resolve the bottleneck.

   Furthermore, this dynamic control extends beyond topology to enable diverse interaction behaviors. We detail these specific patterns, including Vertical Escalation and Lazy Evaluation, in \textbf{Section 3.4 (Control Flow Patterns)}.

\section{Macro-Architecture: The IACT Model}

The macro-architecture of our system is defined by the \textbf{Interactive Agents Call Tree (IACT)}. This section details the topological derivation of the model, its structural mapping to traditional computing, and the specific mechanisms that enable robust multi-agent collaboration.

\subsection{The Derivation of Topology}

To determine the optimal organizational structure, we start from a fundamental constraint: \textbf{The Finite Context Window}. Since complex tasks inevitably exceed the memory of a single model, they must be decomposed into sub-tasks handled by multiple agents with isolated contexts.

However, isolation introduces a critical challenge. Without shared memory, agents naturally diverge in their understanding of the global state and task progress. In this paper, we refer to this phenomenon as \textbf{Agent Decoherence}—analogous to the loss of phase coherence in physics—to describe the \textbf{inevitable divergence in cognitive state among isolated agents}.

This constraint dictates our topological choice:

\begin{itemize}
    \item \textbf{Cyclic Graphs} are rejected because maintaining state coherence across loops requires complex synchronization mechanisms that are prone to failure in probabilistic systems.
    \item \textbf{Linear Chains} are rejected as they lack the capability for hierarchical decomposition and parallel execution.
\end{itemize}

The \textbf{Recursive Tree} emerges as the only non-trivial topology that effectively mitigates decoherence. In a tree structure, a parent agent acts as the exclusive \textbf{router and state keeper} for its subtree. It maintains coherence by strictly controlling the information flow: instructions flow down, and status reports flow up. This ensures that every sub-tree maintains a "Single Source of Truth" rooted in its parent.

\subsection{From Function Tree to Dialogue Tree}

Having established the necessity of the tree topology, we implement it by mapping the components of a traditional Function Call Tree to an agentic system, as shown in Table 1.

\begin{table}[h]
\centering
\caption{Conceptual Mapping between Traditional Computing and IACT}
\begin{tabular}{@{}lll@{}}
\toprule
\textbf{Traditional Computing} & \textbf{IACT Model} & \textbf{Description} \\ \midrule
\textbf{Function} & \textbf{Agent} & An encapsulated computing unit with isolated state. \\
\textbf{Call} & \textbf{Dialogue} & A persistent, stateful interaction session. \\
\textbf{Register/Stack} & \textbf{Context Window} & The limited working memory for immediate processing. \\
\textbf{CPU} & \textbf{LLM} & The probabilistic reasoning engine. \\ \bottomrule
\end{tabular}
\end{table}

In the IACT model:
\begin{enumerate}
    \item \textbf{Encapsulation}: Each agent is treated as an independent function instance with its own isolated memory space (Context Window), preventing information overload.
    \item \textbf{Recursion}: Complex tasks are solved via a divide-and-conquer strategy. An agent can spawn child agents to handle sub-problems, creating a dynamic hierarchy that scales with problem complexity.
    \item \textbf{Stateful Dialogue}: Unlike the instantaneous return of a traditional function call, the interaction between a parent (Caller) and a child (Callee) is a persistent, stateful dialogue. \textit{In a probabilistic system, a command is merely a suggestion that may be misunderstood; thus, the dialogue serves as a necessary mechanism for runtime alignment and error correction.}
\item \textbf{Dynamic Instantiation}: The `CALL` operation serves a dual purpose. On the first invocation, it \textbf{instantiates} a new agent node, effectively spawning a new branch in the topology. On subsequent invocations, it acts as a message channel to the existing instance. This mechanism ensures the topology grows strictly on-demand.
\end{enumerate}

While a standard call tree offers rudimentary fault tolerance—allowing a parent to discard an erroneous result and retry—this mechanism is brittle. Repeated failures often drive the parent agent into hallucination or, more insidiously, lead the agent to subconsciously relax task constraints to achieve a pseudo-success.

IACT addresses this by upgrading the connection to a \textbf{Stateful Dialogue}. This introduces \textbf{Interactional Redundancy}: just as data redundancy enables reliable communication over noisy channels, interactional redundancy enables reliable computation from probabilistic agents.

   The additional communicative turns enable bidirectional correction: the parent can inspect the output and instruct the child to rectify errors, while the child can actively query the parent to resolve ambiguity or request missing context. This mechanism transforms the system from a brittle "retry-based" architecture to a robust "correction-based" architecture.

To sustain coherence across this recursive structure, IACT enforces three fundamental operational principles:
   \begin{enumerate}
       \item \textbf{Principle of Cohesive Decomposition}: The most fundamental mitigation lies in the decomposition strategy itself. Parent agents adhere to the software engineering principle of "High Cohesion, Low Coupling." Tasks are decomposed into sub-problems that are self-contained (e.g., separate modules or sequential stages) rather than tightly coupled logic. By minimizing logical dependencies between siblings, the system reduces the necessity for shared context, rendering decoherence benign.
       \item \textbf{Principle of Exclusive Ownership}: A logical constraint where a persistent resource (e.g., a specific file) should ideally be modified by only one agent during a task. This is maintained by the scheduling agent's autonomous planning.
       \item \textbf{Relevant State Synchronization}: Agents must use dialogue to synchronize their understanding. A parent agent, acting as the coordinator, is responsible for passing only the necessary updated information to children, correcting them if their context appears stale.
   \end{enumerate}

\subsection{Contextual Isolation and Cognitive Efficiency}
A common theoretical concern regarding recursive architectures is the potential for token overhead due to inter-agent communication. However, a deeper analysis reveals that IACT optimizes efficiency by mitigating the "Cognitive Degradation" associated with monolithic contexts.

In flat or shared-context architectures, the context window grows monotonically. While modern inference optimization techniques (e.g., KV Caching) can reduce the computational latency of processing long histories, they cannot prevent the degradation of the model's reasoning capabilities—often referred to as the "Lost in the Middle" phenomenon. As the context fills with irrelevant details from previous sub-tasks, the model's instruction-following ability declines, leading to hallucinations, logical errors, and ultimately, expensive task retries.

IACT addresses this by enforcing strict \textbf{Contextual Isolation}. When a parent agent delegates a task, the child operates in a fresh, clean environment. Upon completion, only the distilled result is returned, while the verbose execution logs remain encapsulated within the child's scope. This ensures that each agent operates within the optimal, high-fidelity region of its context window.

Consequently, IACT prevents the escalating costs associated with error correction in saturated contexts. (For a detailed analysis of the computational efficiency and token economy, see Section 10.3.)

        \subsection{Sequential Execution Model}
        Physically, the current implementation of IACT operates on a \textbf{Sequential Execution Model}. When a parent calls a child, the parent's execution is suspended until the child returns control.

        It is important to note that this single-threaded execution is an implementation choice for simplicity and stability. However, the architecture is inherently compatible with parallel execution. Future extensions can introduce \textbf{non-blocking calls} (e.g., \texttt{CALL\_ASYNC}), allowing a parent to spawn multiple agents simultaneously (e.g., for parallel web research). The results would then be automatically injected into the parent's context or retrieved via an active \texttt{QUERY} mechanism, enabling the system to handle I/O-intensive tasks efficiently.

    \subsection{Topology as Implicit Memory Hierarchy}
    Beyond execution control, the recursive tree serves as a natural \textbf{hierarchical storage management} system. Instead of relying on a monolithic context window, IACT distributes information across the topology based on granularity.

    \begin{itemize}
        \item \textbf{Recursive Information Distillation}: As information flows upward (via `RETURN`), it is progressively compressed from raw execution details into actionable insights. A child agent may process extensive data or perform complex reasoning, but it returns only the synthesized conclusion to its parent. This ensures that higher-level nodes maintain a clean, high-abstraction context.
        \item \textbf{Structural Stability}: The parent node acts as a stable "Anchor" or relative long-term memory for its children. It maintains the broader goal and constraints. If a child agent loses focus due to context saturation, the parent—retaining the original directive—can detect the deviation via the dialogue loop and issue a correction.
    \end{itemize}

    This structure ensures that high-level goals persist at the top of the hierarchy while transient details are encapsulated at the bottom, effectively extending the system's operational context capacity beyond the limits of a single model.

    \subsection{Control Flow Patterns}
IACT's combination of recursive topology and stateful dialogue enables sophisticated control flow patterns that adapt to dynamic problem structures.

\textbf{1. Vertical Escalation (Human-in-the-Loop)}
When a leaf-node agent encounters a decision requiring high-level authorization or subjective judgment (e.g., "Which design style do you prefer?"), it can escalate the query up the agent hierarchy. The request bubbles up through the parent nodes, potentially reaching the User at the root. The User's response then propagates back down, seamlessly integrating human intent into the deep execution loop without breaking the context.

\textbf{2. Dynamic Sibling Delegation}
A sub-agent may realize it lacks specific information or capabilities (e.g., a Coder agent without internet access needs API documentation). Instead of hallucinating, it queries its Parent. Upon receiving the request, the Parent, acting as a dynamic dispatcher, can \textbf{spawn a new sibling agent} (e.g., a Researcher) to acquire the information. Once the Researcher returns the data, the Parent relays it to the Coder, who then resumes execution.

\textbf{3. Lazy Evaluation and Batch Processing}
For tasks generating massive outputs (e.g., "Write a long novel"), returning all content at once would overflow the parent's context window. IACT enables a \textbf{Lazy Evaluation} pattern: the sub-agent generates and returns a segment (e.g., one chapter). The parent processes this segment (e.g., saves it to disk or summarizes it) and requests the next. This iterative "yield-and-process" loop allows the system to handle infinite-length outputs within finite memory constraints.

\textbf{4. Iterative Re-entrancy}
Finally, the IACT topology supports \textbf{Task Re-entrancy}. The agent tree persists after a task cycle completes. A user can inject new requirements into an existing tree (e.g., "Refine the second chapter"). The control flow "re-enters" the established structure, reactivating the relevant sub-agents. This allows for natural, iterative exploration of complex topics without losing the accumulated context or organizational structure.

\section{Micro-Architecture: The Agent Node}

Zooming in from the tree topology, we examine the internal architecture of a single node. The Agent is a sophisticated computing unit driven by a continuous perception-action loop.

\subsection{The Agent Loop}
The operation of an agent can be formalized as a recurrent process. Let $C_t$ be the context at turn $t$, and $A_t$ be the agent's response.
The context is constructed dynamically via a state function $f_{state}$:
\begin{equation}
M_{dynamic} = f_{state}(S_{internal}, S_{external})
\end{equation}
\begin{equation}
C_{t} = (P_{sys}, H_{seq}, I_{in}, M_{dynamic})
\end{equation}
Where:
\begin{itemize}
    \item $P_{sys}$: The static System Prompt defining the agent's role.
    \item $H_{seq}$: The sequence of conversation rounds (history).
    \item $I_{in}$: The input signal for the current turn. This can be a message from the Parent/User (at the start of a turn) or feedback from the Interpreter (during execution).
    \item $M_{dynamic}$: Dynamic system notifications generated by $f_{state}$, such as variable lists, time, or context compression warnings.
\end{itemize}

The LLM generates the response:
\begin{equation}
A_{t} = \text{LLM}(C_{t})
\end{equation}

The Interpreter parses and executes the response to generate new feedback:
\begin{equation}
I_{in}' = \text{Interpreter}(A_{t})
\end{equation}

If $I_{in}'$ is non-empty (e.g., tool output), it is fed back into the next cycle ($C_{t+1}$). If $I_{in}'$ is empty, it indicates that the content of $A_t$ is a message intended for the \textbf{Caller} (Parent or User), and the control flow returns to the upper level.

\begin{figure}[h!]
\centering
\begin{tikzpicture}[node distance=1.5cm]
    \node[basic, fill=gray!5, minimum width=2cm] (context) {Context ($C_t$)};
    \node[basic, fill=blue!10, below right=of context, yshift=-0.5cm] (llm) {LLM ($A_t$)};
    \node[basic, fill=green!10, below left=of context, yshift=-0.5cm] (interp) {Interpreter};
    \node[basic, fill=red!5, below=of interp] (tools) {Tools};

    \draw[arrow] (context) -| (llm);
    \draw[arrow] (llm) -- node[below, font=\tiny] {Natural + Formal} (interp);
    \draw[arrow] (interp) |- node[left, font=\tiny, pos=0.2] {Feedback ($I_{in}$)} (context);
    
    \draw[arrow, <->] (interp) -- node[right, font=\tiny] {Call / Result} (tools);
\end{tikzpicture}
\caption{The Agent Micro-Architecture. The diagram visualizes the cyclic flow between the LLM (Brain) and the Interpreter (Executor). The Interpreter executes the formal code generated by the LLM, and the execution results (Feedback) are fed back to the Context, creating a closed-loop cognitive cycle.}
\end{figure}
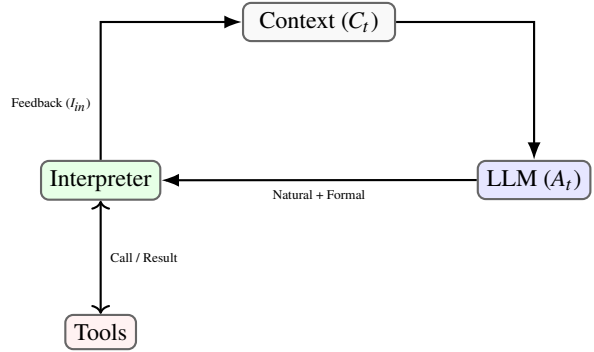

\subsection{Hybrid Language Interpreter}
   The Hybrid Language Interpreter is the core component for implementing text-to-action. It unifies function calls, variable definitions, and references under a single \textbf{pattern-action framework}. The interpreter scans the output stream for registered syntax patterns and maps them to specific executable operations.

   The targets of these actions are unrestricted:
   \begin{itemize}
       \item \textbf{Internal State}: Actions can target the agent's internal state, often influencing its future control flow.
       \begin{itemize}
           \item \textbf{Variable Definition}: The agent can define a variable to store intermediate results. This action adds a new entry to the interpreter's internal variable list, allowing the data to be referenced in subsequent steps.
           \item \textbf{Memory Management}: The agent can invoke \texttt{CONTEXT-COMPRESS}. This action adjusts the agent's internal memory pointer, discarding obsolete history to free up context space.
               \item \textbf{Agent Invocation}: The \texttt{CALL} primitive, responsible for instantiating and communicating with sub-agents, is executed as a standard interpreter action. This unifies agent orchestration with standard tool usage.
        
\end{itemize}
       
       \item \textbf{External World}: Actions can target the external environment. All external tools are encapsulated in \textbf{Ext-Modules}—independent processes that communicate with the core system via RPC. The system dynamically retrieves function lists and documentation from these modules to inject into the context as prompts.
       \begin{itemize}
           \item \textbf{Code Execution}: For instance, to execute code, the agent invokes the \texttt{BASH} function provided by the \texttt{scripter} module. This action triggers an RPC call to the remote module, executing the script in an isolated environment.
       \end{itemize}
   \end{itemize}

\subsection{The Symbolic Variable Mechanism}
A major engineering challenge in LLM systems is the handling of code and structured data. Traditional methods often force LLMs to generate code inside JSON strings, which frequently leads to syntax errors due to escaping issues.

IACT introduces a \textbf{Symbolic Variable Mechanism} to solve this. Variables are defined via internal actions supported by the Interpreter:
\begin{enumerate}
    \item \textbf{Definition}: Variables can be defined in three ways:
    \begin{itemize}
        \item \textbf{Direct Definition}: The agent uses a special syntax to store content.
        \item \textbf{Markdown Capture}: Code blocks in input messages are automatically detected and assigned variable names by the system.
        \item \textbf{Tool/Agent Return}: Outputs from tools (e.g., images) and return messages from sub-agents are automatically wrapped as variables.
    \end{itemize}
    \item \textbf{Propagation}: Variables serve as a reference-based data passing mechanism.
    \begin{itemize}
        \item \textbf{Explicit Argument}: Agents can pass variable names as arguments to tools.
        \item \textbf{Implicit Reference Resolution}: Agents can simply mention a variable name in a message to another agent. The system detects this reference and automatically transports the underlying data (e.g., raw image data) to the recipient's context. This is crucial for passing multimodal content that cannot be easily serialized into a function call string. This mechanism also minimizes token consumption by avoiding data duplication.
        \item \textbf{Distributed Transport}: Since the Core and Ext-Modules may run in different physical environments, local file paths are invalid. Passing variables by value (content) ensures data availability across distributed boundaries.
    \end{itemize}
\end{enumerate}

\subsection{KV Cache Optimization and Context Compression}
To maximize inference speed, the context structure is optimized for \textbf{KV Cache} reuse. Static prefixes (System Prompt, History) are kept at the beginning, while highly dynamic content ($I_{in}$, $M_{dynamic}$) is appended at the end, minimizing the need for cache re-computation.

To maintain efficiency over long horizons, the system implements an active \textbf{Context Compression} mechanism. When context usage exceeds a threshold, the system injects a warning into $M_{dynamic}$. The agent is then responsible for summarizing its history and triggering a truncation operation to free up space.

\subsection{Dynamic Capability Loading}
   Since actions are triggered by syntax patterns, the "Action Space" of an agent is simply the set of patterns it is currently configured to recognize. This set is fluid and managed through three mechanisms:

   \begin{itemize}
       \item \textbf{Static Linking}: Essential patterns (e.g., \texttt{CALL}, \texttt{WAIT}) are hardcoded into the agent's baseline configuration.
       \item \textbf{Contextual Recommendation}: The system continuously analyzes the conversation context. If relevant keywords appear, the system dynamically injects the syntax patterns and documentation of matching tools into the agent's context, enabling "Just-in-Time" tool usage.
       \item \textbf{Runtime Extension}: If a required tool is missing, the agent can load external services (Ext-Modules). The system connects via RPC, retrieves the module's supported syntax patterns, and injects them into the context. Once loaded, these tools become discoverable and usable by \textbf{any agent} in the system whose context matches the tool's description.
   \end{itemize}

\subsection{Just-in-Time Tool Synthesis}
Beyond loading existing modules, IACT possesses the architectural capability to \textbf{synthesize new tools at runtime}. This mechanism allows the system to transform ad-hoc code execution into persistent, reusable capabilities.

When a task requires complex logic, the implementation code can be deployed as a standalone \textbf{Ext-Module process}. The tool functions are subsequently invoked by the Core system via standard RPC channels. This process \textbf{materializes abstract technical knowledge into executable tools}, effectively expanding the agent's action space. By converting verbose script execution into concise function calls, it significantly reduces context pollution and improves the efficiency of subsequent interactions.

\subsection{State-Machine-Based Tooling Paradigm}
To support the principle of Dynamic Context Construction, tools in IACT are designed as \textbf{State Machines}. Instead of exposing a static, exhaustive list of API calls—which imposes a heavy cognitive load—tools expose only the entry points relevant to the current state.

\begin{figure}[h!]
\centering
\begin{tikzpicture}[node distance=0.5cm]
    \node[state] (s1) {\textbf{State 1: Start}\\Prompt: \texttt{Available: BROWSE(url)}};
    \node[state, right=of s1] (s2) {\textbf{State 2: Page Loaded}\\Prompt: \texttt{Content: [Markdown]...}\\ \texttt{Available: SCROLL, CLICK}};
    \node[state, right=of s2] (s3) {\textbf{State 3: Input Filled}\\Prompt: \texttt{Content: [Input]...}\\ \texttt{Available: CLICK(submit)}};
    
    \draw[arrow] (s1) -- node[above, font=\tiny] {BROWSE} (s2);
    \draw[arrow] (s2) -- node[above, font=\tiny] {INPUT} (s3);
    \draw[arrow] (s3) to[bend right] node[above, font=\tiny] {CLICK} (s2);
\end{tikzpicture}
\caption{State-Machine-Based Tooling Paradigm. The diagram illustrates the Browser Tool as a state machine. The available action space (Prompt) changes dynamically based on the tool's state, reducing cognitive load and hallucination.}
\end{figure}
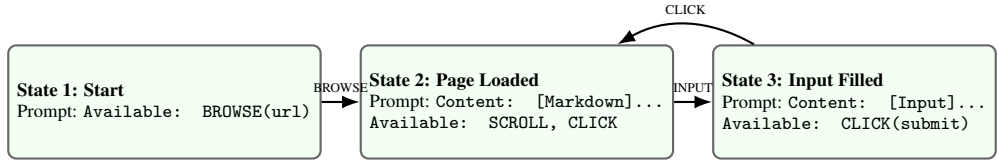

\section{Unified Interaction Protocol}

A collaborative multi-agent system requires a robust communication protocol. IACT adopts a \textbf{Unified Rich-Text Protocol} based on Extended Markdown.

\subsection{Extended Markdown as the Universal Message Exchange Format}
We define Extended Markdown as the sole data exchange format within the system. This choice is primarily driven by the need for \textbf{Multimodal Interoperability}. Since agents (and the user) may possess varying sensory capabilities, the protocol must support the seamless embedding of multimodal data (images, audio, files) alongside text.

The protocol uses a polymorphic embedding syntax: `![Description](Resource\_Identifier)`.
The interpretation of this syntax is context-dependent (\textbf{Protocol Polymorphism}):
\begin{itemize}
    \item \textbf{To a Human User}: The frontend renders it as a UI component (e.g., an image viewer, a file download card).
    \item \textbf{To a Multimodal Agent}: The system resolves the identifier (path or variable) and feeds the content into the multimodal model's input interface.
    \item \textbf{To a Text-Only Agent}: The system preserves the tag as a symbolic reference. The agent can "see" the existence of the file and manipulate it by passing the variable name to other tools (e.g., an OCR tool), even without the sensory capacity to perceive it directly.
\end{itemize}

This design decouples the logical flow of information from the sensory capabilities of individual models, ensuring backward compatibility and seamless collaboration between heterogeneous agents.

\section{Observability and Interactive Supervision}
    The sequential execution model provides a powerful, emergent benefit: \textbf{Observability by Design}.
    \begin{enumerate}
        \item \textbf{Linear, Human-Readable History}: Since agents execute sequentially, their interactions form a single, chronological log. As all messages adhere to the Unified Rich-Text Protocol, this log is not merely text but a rich, renderable history of the entire process, akin to a well-documented email thread.
        \item \textbf{Reconstructible Topology}: The log contains the `CALL` relationships between agents. This allows the linear stream of messages to be programmatically reconstructed into its original tree structure, enabling users to navigate and audit complex workflows by branch and depth.
        \item \textbf{Runtime Intervention}: Building on this observability, the system allows the user to act as a real-time supervisor. At any point, the user can pause execution and inject a message directly into the context of the currently active agent. This provides a mechanism for on-the-fly correction, guidance, or task re-scoping, which is invaluable for high-stakes tasks or when compensating for the limitations of current LLMs.
    \end{enumerate}
    Together, these features form a complete "Observe-Supervise-Intervene" loop, making the agent's complex internal processes transparent and controllable.

    \paragraph{From Breakpoints to Dialogue}
    A critical distinction of IACT is that it operates as a \textbf{General-Purpose Autonomous Agent}. Unlike traditional frameworks where developers must write code to define agent behaviors for specific tasks, IACT requires \textbf{no programming} for task execution. The "program" is simply the user's natural language intent.

    Consequently, "bugs" manifest as cognitive misalignments rather than syntax errors. The "Debugger" is no longer a specialized IDE tool, but the \textbf{Natural Language Interface} itself. Users can pause the execution tree at any depth and inject a clarification (e.g., "You are over-complicating the data cleaning, just use pandas dropna"). This allows users to "patch" the agent's cognitive state in real-time, correcting the execution trajectory without modifying a single line of source code.

\section{Global Associative Memory}

The recursive isolation of IACT, while efficient, introduces a risk of "Tunnel Vision": a leaf-node agent might optimize for a local sub-task while losing sight of the global objective. To mitigate this, we introduce a global associative memory module, termed the \textbf{Hippocampus}.

The Hippocampus operates orthogonally to the call tree. It maintains a vector database of the system's long-term history and global knowledge. During each execution cycle, the system performs a semantic search using the agent's current context as a query. Relevant fragments of global memory—such as the overarching project goal, user preferences, or lessons learned from parallel branches—are dynamically injected into the agent's prompt.

This mechanism balances the benefits of \textbf{Isolation} (efficiency) with \textbf{Coherence} (alignment). It ensures that even a deeply nested sub-agent remains aware of the broader architectural constraints defined at the root level.

\section{System Implementation and Security}

The deployment of autonomous agents capable of executing arbitrary code necessitates a rigorous security posture. We adopt a \textbf{Security by Architecture} approach, operating under a "Zero Trust" assumption regarding the LLM's reliability.

\subsection{Distributed Process Model and Runtime Isolation}
The \textit{kragent.ai} implementation follows a distributed multi-process architecture. The \textbf{Core System} (managing the IACT logic) runs separately from the \textbf{Ext-Modules} (tools and environment).
This separation provides \textbf{Runtime Isolation}. The Core, which handles sensitive logic, runs in a trusted environment. The Ext-Modules, which execute untrusted code (e.g., Python scripts generated by the LLM) or parse untrusted files, are confined within a strict sandbox. This ensures that even if a tool is compromised or crashes, the agent's cognitive state and the core system remain intact.

\subsection{The Sandbox and Root Privileges}
We confront the security paradox: to be useful, an agent needs powerful permissions; to be safe, it must be restricted.
We resolve this by confining agents within a strictly isolated containerized environment where they possess `root` privileges \textit{relative to the container}. This allows them to perform necessary \textbf{Task Execution} (e.g., installing packages, configuring environments) to solve problems. However, the container is strictly isolated from the host system and the internal network. This "Root Inside, Zero Trust Outside" policy maximizes capability without compromising host integrity.

\subsection{Secret Management Paradox}
Agents often need to interact with authenticated external services. However, injecting API keys into the LLM's context is a critical security risk.
We advocate two strategies for \textbf{Keyless Operation}:
\begin{enumerate}
    \item \textbf{Proxy Execution}: For standard integrations, keys are injected into the environment variables of the Ext-Module process, never entering the Agent's context.
    \item \textbf{Ephemeral In-Memory Apps}: For ad-hoc needs, the Agent constructs a temporary Web App. The user inputs secrets directly into this app's UI. The secrets are held in the app's runtime memory (RAM) to perform the task and are wiped upon termination. The LLM orchestrates the app but never "sees" the secret data.
\end{enumerate}

      \section{Practical Lessons and Limitations}
    Our production deployment has revealed several insights and open challenges.

    \subsection{The Multi-Party Dialogue Gap}
    LLMs are primarily trained on two-party dialogue data. In IACT, an agent must interact with three parties: the User/Parent, the System/Tools, and Sub-Agents. Initially, models struggled to distinguish between 'talking to a tool' and 'replying to a parent.' We addressed this by dynamically appending explicit system notes to tool outputs, clarifying their nature as private, system-generated messages invisible to the parent. Notably, recent state-of-the-art LLMs have shown remarkable adaptability to this multi-party context, effectively resolving what was once a significant friction point.

    \subsection{Behavioral Challenges: Passivity under Uncertainty}
    While Decoherence is a structural possibility, its primary driver in practice is a behavioral bias in current LLMs. When facing missing context, models tend to make assumptions and proceed with the task ('hallucinating constraints') rather than pausing to query the parent agent for clarification. This tendency to remain 'silent' exacerbates state divergence, as the system fails to trigger the necessary synchronization dialogues. Future work may involve fine-tuning or RL strategies to encourage proactive questioning.

\subsection{The Memory Bottleneck}
    While the \textbf{Hippocampus} provides long-term retrieval, it lacks temporal awareness. Stale information is often retrieved as relevant fact. Furthermore, semantic search cannot match the associative power of human memory. We envision future iterations incorporating small-scale LLMs dedicated to memory consolidation.

\subsection{Computational Efficiency and Token Economy}
   A valid question regarding recursive architectures is the potential for overhead. However, IACT operates as a \textbf{Lean Architecture}, optimizing efficiency through three mechanisms:

   \begin{itemize}
       \item \textbf{Reduced Prefill Latency}: While inference optimization techniques have improved decoding speeds, processing massive context histories remains computationally intensive. By decomposing a monolithic task into isolated sub-tasks, IACT ensures that each agent operates within a short, focused context window. This minimizes the latency and memory bandwidth consumption associated with prefilling long contexts.
       
        Additionally, the strategy of Dynamic Instruction Injection minimizes the static system prompt, further reducing the baseline token load for every turn.\item \textbf{Zero-Copy Data Transmission}: Traditional agents often re-tokenize massive text blocks (e.g., codebases, long documents) when passing them between steps. IACT employs the \textbf{Symbolic Variable Mechanism} (Section 4.3) to exchange lightweight references rather than raw data. This ensures that communication overhead remains constant ($O(1)$) regardless of the content size.
       
       \item \textbf{Amortized Structural Overhead}: As noted in our empirical observations, for the complex, long-horizon tasks IACT is designed to solve (often exceeding $100k$ tokens), the structural overhead is negligible. In such scenarios, a typical workflow might involve fewer than a dozen agent instantiations. The communication between agents—the "Call" and "Return" messages—primarily consists of control signals and high-level summaries. Bulk data is typically exchanged via file paths or the Symbolic Variable Mechanism, bypassing the message stream. As a result, the total token consumption for structural management typically remains in the low thousands, a fraction of the total workload. While the latency of the recursive interaction is perceptible in trivial, short-turn interactions, it is effectively amortized to zero in substantial engineering tasks.
   \end{itemize}

\section{Case Studies: Real-World Engineering \& Research}
To demonstrate the system's capability in handling open-ended tasks, we present two unedited execution records from early 2024. These cases demonstrate how IACT enables models (e.g., GPT-4 class) to perform complex engineering and research tasks through \textbf{iterative, multi-turn collaboration}.

\subsection{Autonomous Service Deployment (April 1, 2024)}
\textbf{Video Record:} \href{https://youtu.be/fLGR2iiXviA?si=T8MxVpTrwSFV4GAW}{Watch on YouTube}

\textbf{The Workflow:}
This task was not completed via a single prompt but through a progressive three-stage evolution, demonstrating the system's ability to maintain context and iterate on code.
\begin{enumerate}
    \item \textbf{Phase 1: Market Research}: The user asked the agent to identify popular text-to-image models on Hugging Face. The agent returned a list of candidates.
    \item \textbf{Phase 2: Proof of Concept (PoC)}: The user selected a model and instructed the agent to generate a specific test image ("a fat orange cat"). The agent autonomously located the model's documentation, wrote a Python script to load weights, and successfully generated the image locally.
    \item \textbf{Phase 3: Service Productization}: Leveraging the code from the PoC phase, the user requested a web interface. The agent refactored the inference script into a Flask application, handled the frontend logic, and launched a local web service (127.0.0.1:59012), allowing the user to generate images via a browser.
\end{enumerate}

\subsection{Multimodal Knowledge Extraction (May 26, 2024)}
\textbf{Video Record:} \href{https://youtu.be/8OqBgJE6ETQ?si=heLXHL3Dj9a5GFFX}{Watch on YouTube}

\textbf{The Workflow:}
This case illustrates a complex \textbf{Tool Chaining} pipeline and the system's responsiveness to human feedback.
\begin{enumerate}
    \item \textbf{Phase 1: Data Acquisition}: The user instructed the agent to download Richard Feynman's lecture videos from YouTube. The agent handled file system operations to organize the data.
    \item \textbf{Phase 2: Audio Processing \& Correction}: The agent initially extracted audio from standard video files. The user noticed that `.webm` files were skipped and issued a correction ("There are many videos in webm format..."). The agent acknowledged the oversight, modified its script to handle the specific container format, and successfully processed the remaining files.
    \item \textbf{Phase 3: Transcription Pipeline}: The agent identified the `whisper-large-v3` model on Hugging Face, studied its usage, and wrote a script to transcribe the audio files into a merged text document.
    \item \textbf{Phase 4: Reasoning over Data}: Finally, the user asked a specific physics question ("Why do we need antiparticles?"). The agent retrieved the answer solely from the transcribed text, synthesizing Feynman's explanation.
\end{enumerate}

\subsection{Empirical Conclusion}
We deliberately present case studies using historical models (early 2024) to address a common misconception: that robust autonomous agents await future "super-models." Our experience confirms that the architecture is compatible with a range of models. Historically, even early open-source models like Mistral 7B demonstrated capability on simpler tasks. Today, the system achieves peak performance with state-of-the-art proprietary models (e.g., Claude series, Gemini series). Leading open-source models (e.g., GLM and Qwen series) are also compatible and viable, though generally performing a tier below in complex execution tasks.

   It is crucial to note that \textbf{raw intelligence alone is insufficient}. Our experiments indicate that some recent high-performing models fail to execute IACT workflows. This is likely due to aggressive fine-tuning for proprietary function-calling schemas, which interferes with the system's native text-to-action protocol.

   Thus, IACT acts as a \textbf{structural scaffolding} that \textbf{guides and releases} the potential of \textbf{compatible} models. This validates that the architecture is viable \textit{today} with proper model selection, unlocking complex behaviors without waiting for next-generation reasoning engines.

\section{Related Work}
   The development of IACT is situated within the landscape of autonomous agents but diverges fundamentally in its architectural focus. We distinguish IACT from existing paradigms along three dimensions: topology, execution model, and system abstraction.

   \subsection{Cognitive Search vs. Organizational Structure}
   Research such as \textbf{Chain-of-Thought (CoT)} \cite{wei2022chain} and \textbf{Tree of Thoughts (ToT)} \cite{yao2023tree} explores the internal reasoning topology of a single model. ToT, for instance, utilizes a tree structure as a \textit{search algorithm} to explore and backtrack through reasoning paths.
   In contrast, IACT employs the tree structure not for cognitive search, but as an \textbf{organizational architecture} for task execution. Unlike ToT, which occurs within a single context window, IACT's tree spans multiple isolated agent instances. It addresses the engineering challenge of scaling context and managing state across isolated agent instances.

   \subsection{Static Graphs vs. Dynamic Topology}
   Multi-agent frameworks like \textbf{CAMEL} \cite{li2023camel} and \textbf{AutoGen} \cite{wu2024autogen} pioneered the paradigm of "Communicative Agents," typically modeling collaboration as a conversation between roles. These frameworks often rely on pre-defined interaction graphs or static workflows.
   IACT advances this by introducing a strictly \textbf{LLM-Centric Control Flow} that manifests as a \textbf{Dynamic Topology}. Unlike frameworks that often require pre-configured graphs, IACT operates as a \textbf{general-purpose autonomous agent} capable of executing open-ended tasks directly from raw user input. The organizational structure is not fixed by the developer but emerges incrementally at runtime. Agents autonomously spawn sub-agents and extend the hierarchy only when the task complexity demands it, allowing the system structure to adapt fluidly to open-ended problems.

   \subsection{Embodied Learning vs. General-Purpose Runtime}
   \textbf{Voyager} \cite{wang2024voyager} demonstrated the power of lifelong learning in embodied environments (Minecraft) by generating and retrieving executable code skills.
   IACT generalizes this "Code-as-Skill" philosophy into a \textbf{General-Purpose Runtime}. Instead of being bound to a specific game engine, IACT agents operate within a \textbf{Pattern-Action Interpreter} capable of interacting with the OS and external services. This allows agents to not only generate scripts but to dynamically construct and load new \textbf{Ext-Modules}, effectively expanding the system's capability set for any domain, from software engineering to data science.

\section{Conclusion}

We have presented the \textbf{Interactive Agents Call Tree (IACT)}, a computational model that effectively addresses the limitations of static topologies inhibiting current agent systems. By abandoning pre-defined workflows in favor of a \textbf{self-organizing, recursive hierarchy}, IACT transforms the agent from a rigid tool into a dynamic entity capable of adapting to open-ended complexity.

Through mechanisms like \textbf{Contextual Isolation}, \textbf{Cohesive Decomposition}, and \textbf{Stateful Dialogue}, IACT provides a resilient framework where raw reasoning power is channeled into productive, self-correcting workflows. This architecture demonstrates that with the right organizational principles, existing high-capability models can already bridge the gap between probabilistic reasoning and reliable engineering.

As we move towards General AI, the focus must shift from designing static scripts to cultivating dynamic systems. IACT offers a blueprint for this transition: an architecture that does not merely execute plans, but \textbf{grows} them. It stands as a foundational step towards fully autonomous systems that can navigate the ambiguity of the real world with the rigor of software engineering.

\section*{System Availability}

The architecture described in this paper is fully implemented and deployed at \textbf{\href{https://kragent.ai}{kragent.ai}}.

Designed as a \textbf{Research-First Execution Environment}, the platform leverages the IACT model to automate complex academic and engineering workflows. It provides researchers with an autonomous assistant capable of:

\begin{itemize}
    \item \textbf{Autonomous Software Engineering}: Building and deploying full-stack web applications, interactive 3D simulations, and performing static analysis on existing code repositories.
    \item \textbf{Computational Science \& Data Pipelines}: Executing end-to-end data workflows, from retrieving raw data (e.g., PDB, NCBI, Financial Reports) to training ML models and generating numerical simulations.
    \item \textbf{Deep Research \& Data Collection}: Conducting long-horizon investigations across medical, legal, and technical domains, \textbf{automating the collection and organization of datasets} (e.g., tutorials, case laws) for domain-specific analysis.
    \item \textbf{Academic Production}: Autonomously drafting scientific papers in \LaTeX, converting citation formats, and compiling camera-ready PDFs and presentation slides.
\end{itemize}

We invite the research community to utilize this platform to explore the boundaries of Agentic Computing.

\end{document}